\documentclass[letterpaper, 10 pt, journal, twoside]{IEEEtran}
\usepackage{cite}
\usepackage{array}
\usepackage{url}

\usepackage{graphics} 
\usepackage{epsfig} 
\usepackage{amsmath} 
\usepackage{amssymb}  
\usepackage{xspace}
\usepackage{bm}

\usepackage{algorithm}
\usepackage{algpseudocode}

\usepackage{pgfplots}
\usepackage{tikz}
\usepackage{graphicx}
\usepackage[caption=false,font=footnotesize]{subfig}  
\pgfplotsset{compat=1.16}
\usepackage{tabularx}

\usepackage{booktabs}
\usepackage{xcolor}

\usepackage[acronym]{glossaries}

\newacronym{hydo}{HyDo}{Hybrid Diffusion Policy}
\newacronym{hacman}{HACMan}{HACMan}

\newcommand{\cc}[1]{{#1}}
\newcommand{\ploc}{\pi^{\rm{loc}}}
\newcommand{\pmot}{\pi^{\rm{m}}}
\newcommand{\rebuttal}[1]{#1}

\newcommand{\ba}{\bm{a}}
\newcommand{\bs}{\bm{s}}

\begin{document}
%
\title{Enhancing Exploration with Diffusion Policies in Hybrid Off-Policy RL: Application to Non-Prehensile Manipulation}
%
%
%

\author{Huy Le, Tai Hoang, Miroslav Gabriel, Gerhard Neumann, Ngo Anh Vien%
\thanks{Huy Le is with the Bosch Center for Artificial Intelligence, Renningen, Germany and also with the Institute for Anthropomatics and Robotics, Karlsruhe Institute of
Technology, 76131 Karlsruhe, Germany.
        {\tt\small baohuy.le@de.bosch.com}
        }%
\thanks{Miroslav Gabriel, Ngo Anh Vien are with the Bosch Center for Artificial Intelligence, Renningen 71272, Germany
        }%
\thanks{Tai Hoang, Gerhard Neumann are
with the Institute for Anthropomatics and Robotics, Karlsruhe Institute of
Technology, 76131 Karlsruhe, Germany.
        }%
}
\maketitle

\begin{abstract}
Learning diverse policies for non-prehensile manipulation is essential for improving skill transfer and generalization to out-of-distribution scenarios. In this work, we enhance exploration through a two-fold approach within a hybrid framework that tackles both discrete and continuous action spaces. First, we model the continuous motion parameter policy as a diffusion model, and second, we incorporate this into a maximum entropy reinforcement learning framework that unifies both the discrete and continuous components. The discrete action space, such as contact point selection, is optimized through Q-value function maximization, while the continuous part is guided by a diffusion-based policy. This hybrid approach leads to a principled objective, where the maximum entropy term is derived as a lower bound using structured variational inference. We propose the Hybrid Diffusion Policy algorithm ({\bf \acrshort{hydo}}) and evaluate its performance on both simulation and zero-shot sim2real tasks. Our results show that HyDo encourages more diverse behavior policies, leading to significantly improved success rates across tasks - for example, increasing from 53\% to 72\% on a real-world 6D pose alignment task. Project page is available at \url{https://leh2rng.github.io/hydo}
\end{abstract}


%
\IEEEpeerreviewmaketitle

\section{Introduction}
%
%
%
%
The ability to manipulate objects in ways beyond simple grasping is a vital aspect of human dexterity, underscoring the significance of learning advanced non-prehensile manipulation skills. These complex skills are essential for a wide range of tasks, from daily activities to advanced industrial applications. Teaching robots to achieve a level of dexterity similar to the one of humans remains a significant challenge for the field of robotics \cite{yu2016more,xu2022universal}.
Previous research has made significant advances in this area, but often suffers from limitations in object generalization and motion complexity \cite{cheng2022contact,hou2019robust,mo2021where2act}. To address these challenges, motion primitives (MPs) are frequently employed to simplify the representation of long-horizon actions and thus the overall problem complexity. In addition, object-centric action representations are utilized to decrease the sample complexity and to enable a more efficient learning process.
Reinforcement Learning (RL) can be used to learn such representations, especially within hybrid action spaces that combine discrete contact points with continuous parameters for MPs \cite{FeldmanZVC22,ZhouJYPH23}.

Developing policies that can learn diverse behaviors in a RL context is motivated by two key factors: First, these policies are potentially able to improve generalization to out of distribution states and observations. Policies that only overfit to a narrow range of experiences, on the other hand, usually do not perform sufficiently well in unseen environments. Continually learning from a more diverse set of experiences forces the policy to capture the underlying principles of the task rather than merely optimizing for specific scenarios encountered during training \cite{merel2018neural,kumar2020one}.
Secondly, developing such policies enhances skill transfer learning: Continuous exposure to diverse experiences in online RL facilitates the transfer of skills across different but related tasks. An agent that learns from a broad spectrum of interactions is more likely to develop a robust set of skills that can be applied to a varity of tasks. This increases its versatility and overall learning capability \cite{merel2018neural,kumar2020one,jia2024towards}.

\begin{figure*}[ht]
    \centering
    \includegraphics[width=0.85\linewidth]{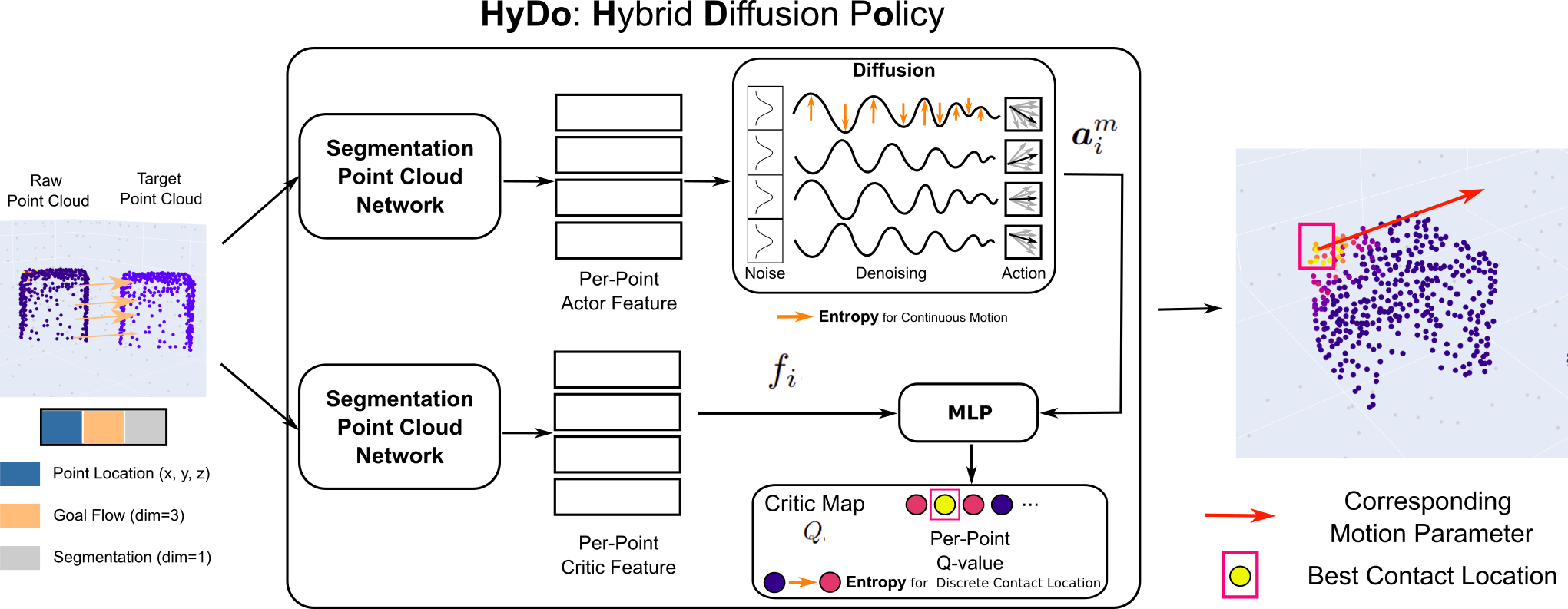}
    \caption{Overview of HyDo: The network takes point clouds, goal flow, and segmentation (indicating object and background points) as input. These are passed through the actor and critic networks. The actor is enhanced exploration on the continuous motion parameter with the entropy regularizer applied during the diffusion process and outputs the motion parameter. The state-action pair is then evaluated by the critic, which also integrates entropy regularization for exploration on the discrete contact location. The action parameter and contact location with the highest Q-value is selected and executed by the robot.}
    \label{fig:diagram}
\end{figure*}

In this work, we introduce a novel approach to enhancing exploration in non-prehensile manipulation tasks under a hybrid off-policy reinforcement learning framework \cite{ZhouJYPH23}. Our method handles both discrete and continuous action spaces by incorporating maximum entropy principles to encourage diverse behaviors. Specifically, we represent the continuous motion parameter policy using a diffusion model \cite{song2019generative, ho2020denoising, ding2023consistency, chi2023diffusion}, while the discrete action space, such as contact points, is optimized through Q-value function maximization. This formulation leads to the development of a Hybrid Diffusion Policy algorithm, called {\bf \acrshort{hydo}}, which integrates two main components: diffusion-based policies and maximum entropy optimization over both discrete and continuous actions. The entropy maximization term, embedded in the soft actor-critic (SAC) \cite{haarnoja2018soft} algorithm, is derived as a lower bound using structured variational inference. The overall framework is illustrated in Fig.~\ref{fig:diagram}. We evaluate the impact of combining maximum entropy regularization with diffusion in both simulation and zero-shot sim2real tasks. The results show that this combination helps to learn more diverse behavior policies. In the zero-shot sim2real transfer, this improved exploration leads to a significant increase in success rates, from 53\% to 72\% on a 6D pose alignment task with a physical Franka Panda robot.

In summary, our contributions are: i) a hybrid RL framework that enhances exploration by incorporating diffusion-based policies; ii) the integration of maximum entropy regularization to encourage diverse behaviors across both action spaces; and iii) a theoretical justification, showing that the new objective is a lower bound derived via structured variational inference. We validate our methods on both simulated and zero-shot sim2real non-prehensile manipulation tasks.
%

\section{Related work}
\label{sec:related_work}

\subsection{Diffusion-based offline and online RL}

There have been different efforts, such as diffusion-based generative models DDPMs \cite{ho2020denoising}, to use diffusion models as representations for RL policies because they are (among others) capable of  learning multi-modal and diverse behaviors. Most works focus on offline RL in which all data is available at training time. Diffusion-Q learning \cite{wang2022diffusion} proposes an explicit regularization of the cloned behavior policy. A follow-up work \cite{ding2023consistency} proposes to use consistency models as policies since they improve inference speeds such that they can be applied to online RL. Other variants of diffusion policies introduce actor-critic methods via implicit Q-learning \cite{hansen2023idql}, Q-guided
policy optimization with a new formulation for intermediate guidance in diffusion sampling process \cite{lu2023contrastive}, diffusion-constrained Q-learning on latent spaces \cite{venkatraman2023reasoning} or implicit Q-learning as an actor-critic method \cite{hansen2023idql, li2024learning}.

Analogically, diffusion policies are  applied in imitation learning \cite{chi2023diffusion}, goal-conditioned imitation learning \cite{reuss2023goal}, human behavior imitation learning \cite{pearce2023imitating}, offline direct policy search \cite{chen2022offline,he2023diffcps} based on advantage-weighted regression (AWR) or reward-weighted regression (RWR) \cite{peters2007reinforcement}.
There are also few works which apply online RL to fine-tune pretrained diffusion models, such as RWR-based methods \cite{lee2023aligning,black2023training} or advanced bandit setting-based methods \cite{Uehara2024,abs-2402-15194}. While these RWR-and AWR-based methods are considered to implicitly enforce entropy-max regularization, they are only 1-step MDPs and ignore the exploration problem for training from scratch. Closest to our work is Q-score matching for actor-critic (QSM) \cite{psenka2023learning} which applies soft policy iteration, but without direct use of the soft $Q$-value function. Our work however optimizes the entropy-max objective using soft Q-value functions for the policy improvement step as well as the soft Bellman updates of the policy evaluations step.

\subsection{Manipulation skill learning with motion primitives}

Non-prehensile manipulation involves manipulating objects without grasping them.
Recent learning-based methods in this field are limited either by skill complexity or skill diversity \cite{mo2021where2act,xu2022universal,zhou2023learning}. Our work tries to address these challenges by focusing on learning diverse behaviors for 6D object manipulation with MPs and by optimizing a class of multi-modal policies. Traditional RL algorithms focus either on discrete or on continuous action spaces, despite the fact that some applications require hybrid action spaces where the agent selects a discrete action along with some continuous parameter. Recent approaches \cite{FeldmanZVC22,ZhouJYPH23,jiang2024hacmanpp} address these shortcomings by using a spatial action representation with discrete actions defined over a visual input map but fail to incorporated 
profound exploration strategies because they either ignore exploration over spatial maps \cite{FeldmanZVC22} or use a simple $\epsilon$-greedy strategy to explore over a continuous action parameter space \cite{ZhouJYPH23}. Another recent work proposes a diffusion-based MP policy \cite{Scheikl2023} that is similar to us that the policy generates only MP parameters which are then used to generate a full motion profile via ProDMP \cite{li2023prodmp}.

\section{Background}

\subsection{MDP and Off-Policy Actor-Critic Methods} 
The underlying problem of RL can be formulated as a Markov decision process (MDP) \cite{sutton2018reinforcement} which is defined by a tuple $\{\cc{S}, \cc{A}, \cc{R}, \cc{P},\gamma, \cc{P}_0\}$ with state space $\cc{S}$, action space $\cc{A}$, rewards $\cc{R}$, transition probabilities $\cc{P}$, discount factor $\gamma$, and initial state distribution $\cc{P}_0$.
The objective is to maximize the discounted cumulative reward $R = \sum_{i=0}^{\infty} \gamma^i r(s_{t+i},a_{t+i})$ with states $s_t$ and actions $a_t$.
Off-policy actor-critic methods decouple the policy used to generate data from the policy being optimized. The \textit{actor} is represented by a policy $\pi_{\theta}(a|s)$ with parameters $\theta$, whereas the \textit{critic} estimates the value function $Q_{\phi}(s, a) = \mathbb{E}\left[\sum_t \gamma^t r(s_t,a_t)|s_0=s, a_0=a\right]$ with parameters $\phi$. 
TD3 \cite{fujimoto2018addressing} and SAC \cite{haarnoja2018soft} are two prominent off-policy approaches. The objectives of SAC's actor and critic are defined as follows: 


\begin{align}
\begin{split}
L(\theta) = \mathbb{E}_{s,a \sim \cc{D}} \left[ -Q_\phi(s, a) + \alpha \log \pi_\theta(a|s) \right], \\
L(\phi) = \frac{1}{2} \mathbb{E}_{s_t,a_t,s_{t+1} \sim \cc{D}} \left[ \|Q_{\phi}(s_t,a_t) - y_t\|^2 \right],
\label{critic-loss}
\end{split}
\end{align}
where $y_t$ is the target value, described as
\begin{align*}
y_t = r_t + \gamma \, \mathbb{E}_{a_{t+1} \sim \pi_{\theta}(s_{t+1})} \big[
    & Q_{\phi'}(s_{t+1}, a_{t+1}) \notag  \\ 
    & - \alpha \log \pi(a_{t+1}|s_{t+1}) \big].
\end{align*}

In contrast to TD3 that optimizes deterministic policies and exploration is handled by an $\epsilon$-greedy strategy, SAC uses a stochastic $\pi_{\theta}$ and adds an entropy maximization term to the objective to encourage exploration. This promotes a more diverse set of behaviors and provides an improvement in exploration and training stability.

\subsection{Diffusion and Consistency Models}
\label{sec:intro_diff_vs_cm}

Diffusion-based generative models DDPMs \cite{ho2020denoising,song2019generative} assume \( p_\theta(x_0) := \int p_\theta(x_{0:T}) \, dx_{1:T} \), where \( x_1, \ldots, x_T \) are latent variables with the same dimensionality as the data \( x_0 \sim p(x_0) \). In a forward diffusion chain \(q\), given by $q(x_{1:T} | x_0) := \prod_{t=1}^{T} q(x_t | x_{t-1})$,  $q(x_t | x_{t-1}) := \mathcal{N}(x_t; \sqrt{1 - \beta_t} x_{t-1}, \beta_t I)$,
noise with a predefined variance schedule \( \beta_i \) is gradually added to the data over a fixed amount of time steps \( T \).
Subsequently, a reverse diffusion chain \(p\), defined as $
    p_\theta(x_{0:T}) := \mathcal{N}(x_T; 0, I) \prod_{t=1}^{T} p_\theta(x_{t-1} | x_t), $
is optimized by maximizing the evidence lower bound. Inference then requires sampling the reverse diffusion chain from \( t = T \) to \( t = 0 \).


\rebuttal{However, diffusion models rely on progressive denoising over a large number of steps, which can lead to slow sampling speed. This can be a critical bottleneck for online RL where it heavily depends on sampling from environments}. Consistency models \cite{song2023consistency} extend diffusion models by adopting the form of a probability flow ordinary differential equation (ODE). The reverse process along the ODE path \(\{\hat{x}_\tau\}_{\tau \in [\epsilon, T]}\) generates data starting from \(\hat{x}_T \sim \mathcal{N}(0, T^2 I)\), where \(\epsilon\) is a small value close to zero. Consistency models learn a parameterized function $f_{\theta} : (x_{\tau}, \tau) \rightarrow x_{\epsilon}$ which maps from a noisy sample $x_{\tau}$ at step $\tau$ to its original sample $x_{\epsilon}$. \rebuttal{Furthermore, as detailed in Section~\ref{sec:diff_vs_consistency}, our experiments show that, despite a modest speedup, consistency models empirically achieve a higher success rate than pure diffusion methods.}

\subsection{Diffusion and Consistency Models as RL Policy}

Diffusion models have been used as a new class of policies in offline RL \cite{wang2022diffusion,kang2024efficient} for actor-critic architectures. These works share a similar parametric policy representation as the reserve process of the conditional diffusion model which is defined as
\begin{align}
    \pi_\theta(\bm{a}|\bm{s}):=\pi_\theta(\bm{a}^{0:K}|\bm{s}) =\mathcal{N}(\bm{a}^K;\bm{0},\bm{I}) \prod_{k=1}^K p_\theta(\ba^{k-1}|\ba^k,\bs)
\label{diffusion_policy}
\end{align}
where $\ba^0$ is the action executed by the agent and sampled at step 0.
The probability distribution 
$p_\theta$ can be based on DDPM \cite{wang2022diffusion,kang2024efficient} in which $p_\theta(\ba^{k-1}|\ba^k,\bs)$ is a Gaussian with a trainable mean $\mu(\ba^k;k,\bs)$ and a fixed time-dependent covariance $\Sigma(\ba^k;k,\bs)=\beta^k \bm{I}$.

With the above policy representation, Wang et. al. \cite{wang2022diffusion} propose Diffusion Q-learning to optimize $\theta$ using a similar update scheme as TD3+BC \cite{fujimoto2021minimalist}. Given an offline dataset $\cc{D}=\{\bs_t,\ba^0_t,r_t,\bs_{t+1}\}_{t=0:T}$, the objective of the policy evaluation step is
\begin{align*}
L(\phi) = \mathbb{E}_{\bs_t, \ba^0_t, r_t \sim \cc{D}} \left[ \sum_{i=1}^2 \|y_t - Q_{\phi_i}(\bs_t,\ba^0_t)\|^2 \right],
\end{align*}
where $y_t=r(\bs_t, \ba^0_t) +\gamma \min_{i=1,2}Q_{\phi'_i}(\bs_{t+1},\ba^0_{t+1})$, 
$\ba^0_{t+1}\sim \pi_\theta(\ba|\bs)$, and $\{Q_{\phi_i}\}_{i=1,2}$ are twin Q networks with parameters $\phi=\{\phi_1,\phi_2\}$ and with target networks $Q_{\phi'_i}$. The objective of the policy improvement step is $ L(\theta) = L_{\rm{behavior\_cloning}}(\theta) - \alpha \mathbb{E}_{\bs \sim \cc{D}, \ba^0 \sim \pi_\theta}\left[Q_{\phi_1}(\bs,\ba^0) \right]$,
where $\alpha$ is the trade-off hyperparameter between two losses. The behavior cloning loss is a standard supervised learning loss of DDPM, i.e. fitting a diffusion prediction model $\pi_\theta(\cdot|\bs)$ to predict $\ba^0$.

The main drawback of optimizing the objectives $L(\phi)$ and $L(\theta)$ is the high computational demand of having to perform a multitude of sampling steps in order to obtain $\ba^0 \sim \pi_\theta(\cdot|\bs)$.
Ding et. al. \cite{ding2023consistency} propose using consistency models $f_\theta(\bs,\ba^\tau,\tau)$ with $\pi_\theta(\bs)={\texttt{Consistency\_Inference}}(\bs,f_\theta)$ \cite{song2023consistency} to reduce the number of steps and show that they can effectively be applied for online RL.

\subsection{Hybrid Actor-Critic for Non-prehensile Manipulation}

Our work is based on Feldman et al. \cite{FeldmanZVC22} and HACMan \cite{ZhouJYPH23}. Both propose to use a hybrid action space which consists of a continuous action space for motion prediction and a discrete action space for inferring contact locations, and employ similar actor-critic based network architectures. 
They learn an actor which predicts a per-point motion parameter map $\ba^m = \{\ba^m_i = \pi_\theta(\cc{X}) \mid i=1\ldots N \}$ for a given input point cloud $\cc{X}$, as well as a critic which determines a per-point Q-value map $Q=\{Q_i=Q_\phi(\cc{X},\ba^m_i) \mid i=1\ldots N\}$ for the motion parameter of each point $\ba^m_i$. Both networks share a common encoder $f(\cc{X})=\{f_i \mid i=1\ldots N\}$ which predicts a per-point feature map. Based on $Q$, a location policy $\pi_{loc}$ then selects a discrete point $x_i$ and, in that way, also the corresponding continuous $\ba^m_i$. Feldman et. al. \cite{FeldmanZVC22} add a max-entropy term for exploring the continuous motion prediction space, but ignore the exploration on discrete location space. HACMan \cite{ZhouJYPH23} uses TD3 which resorts to a simple $\epsilon$-greedy strategy for exploring both locations and motion parameters. It computes the probability of a point being selected as the contact location by 
\begin{align}
    \pi_{\rm{loc}}(x_i \mid \cc{X_{obj}}) = \frac{\exp(\beta Q_i)}{\sum_{k=1..N} \exp(\beta Q_k)},
    \label{loc_policy}
\end{align}
where $\beta$ is the softmax's temperature, and $N$ is the number of points on the object point cloud $\cc{X}_{obj}$\footnote{Points on the background point cloud $\cc{X}_{b}$ determined through separate segmentation component are not considered.}.

\section{Methodology}

\subsection{Problem Formulation}


The goal of this work is to develop policies for non-prehensile manipulation tasks, specifically targeting 6D object pose alignment. This task requires the policy to process a point cloud \(X\) as input, where each point is represented by its 3D coordinates, a 1D segmentation mask, and a 3D goal flow vector. To solve this problem, the policy must handle both discrete actions, such as selecting contact points, and continuous actions, like generating motion primitive vectors.

We formulate this problem in a principled manner as an online off-policy maximum entropy reinforcement learning task. This framework is chosen to encourage exploration and diversity in the learned behaviors. To represent the policies, we leverage Denoising Diffusion Probabilistic Models (DDPMs) \cite{ho2020denoising} and Consistency Models (CMs) \cite{song2023consistency}, which are well-suited for capturing diverse behaviors. Specifically, we first introduce a principled formulation to integrate diffusion policies into the Soft Actor-Critic (SAC) algorithm, enabling the continuous policies to capture multi-modalities. We then extend this approach to support both discrete and continuous actions within a hybrid RL framework, leading to our Hybrid Diffusion Policy (HyDo). This principled formulation enhances exploration across both action spaces, ensuring the development of diverse and robust manipulation policies.



\subsection{Soft Actor-Critic with Diffusion Policy}
\label{sac-diffusion}

Given a policy $\pi$ parameterized by a diffusion model defined in Eq.~\ref{diffusion_policy}, we propose to incorporate a Diffusion Policy that optimizes an objective similar to SAC, i.e. the entropy-regularized cumulative return:
\begin{align}
\begin{split}
    J_\pi(\theta) = \sum_t  \mathbb{E}_{\bs_t,\ba^{0:K}_t \sim\pi_\theta} \Bigg [ &r(\bs_t,\ba^0_t) \\
    &- \alpha\sum_{k=0}^K \log \pi_\theta(\ba^{k-1}_t|\ba^{k}_t, k, \bs_t) \Bigg ],
\label{soft-return}
\end{split}
\end{align}
where $\alpha$ is a hyperparameter. The entropy term in Eq.~\ref{soft-return} can also be interpreted as $-\log p(\ba_t|\bs_t)$ of the whole sampling action diffusion path instead of only $-\log p(\ba^0_t|\bs_t)$ (with the true RL action) as in the standard SAC's objective, because it is intracble to compute the density of diffusion models. We follow the derivation of the structured variational inference
\cite{levine2018reinforcement} to prove that $J_\pi(\theta)$ is the lower-bound of the maximum reward likelihood,
\begin{align*}
    \log p (\cc{O}_{1:T}) &\geq \mathbb{E}_{\bs_{1:T}, \ba^{0:K}_1,\ldots, \ba^{0:K}_T \sim q(\bs_{1:T}, \ba^{0:K}_1,\ldots, \ba^{0:K}_T) } \nonumber \\
    &\quad \left[ \sum_{t=1}^T r(\bs_t,\ba^0_t) - \alpha \sum_{k=0}^K \log q(\ba^{k-1}_t|\ba^{k}_t,k,\bs_t) \right],
\end{align*}
where the binary random variable $\cc{O}$ denotes if time step $t$ is optimal or not, and $q$ is the variational distribution, in which the distribution over $\cc{O}$ is $p(\cc{O}_t|\bs_t,\ba_t) = \exp(\frac{1}{\alpha}r(\bs_t,\ba_t))$. The proof applies  Jensen's inequality as follows,
\begin{align*}
    \log & p (\cc{O}_{1:T}) =\log \int p (\cc{O}_{1:T},\tau) d \tau\\
     &\ge \mathbb{E}_{\tau \sim q(\tau)} \left[\log p (\cc{O}_{1:T},\tau) - \log q(\tau) \right] \\
     &= \mathbb{E}_{\tau \sim q(\tau)} \left[\sum_{t=1}^T \left( r(\bs_t,\ba_t) -\alpha \sum_{k=1}^{K} \log q(\ba_{t}^{k-1}|\ba_{t}^{k},\bs_t) \right) \right], \\
\end{align*}
where the variational policy distribution $q(\ba_{t}^{k-1}|\ba_{t}^{k},\bs_t)$ is parameterized as $\pi_\theta(\ba_{t}^{k-1}|\ba_{t}^{k},k,\bs_t)$.

As a result, we obtain the policy evaluation step of the soft policy iteration with the modified soft Bellman backup operator $\cc{T}^\pi$
\begin{align}
    \cc{T}^\pi Q(\bs_t,\ba^0_t) &= r(\bs_t,\ba^0_t) \nonumber + \gamma \mathbb{E}_{\ba_{t+1}^{0:K} \sim \pi, \bs_{t+1}\sim\cc{P}} \Bigg[ Q(\bs_{t+1}, \ba^0_{t+1}) \nonumber \\
    &\quad - \alpha \sum_{k=0}^K \log p_\theta(\ba^{k-1}_{t+1}|\ba^{k}_{t+1},k,\bs_{t+1}) \Bigg].
\end{align}

The policy improvement step updates the policy with the same objective as SAC, i.e., we minimize $
   \cc{L}(\theta) = \mathbb{E}_{\bs_t \sim \cc{D}} \left[ {\rm D}_{{\rm KL}} \left( \pi'(\cdot|\bs_t) \Vert \frac{\exp(Q_\phi(\bs_t,\cdot))}{Z_\phi(\bs_t)}\right) \right].
$
Using DDPMs or CMs \cite{ding2023consistency}, each probability $p_\theta(\ba^{k-1}_{t}|\ba^{k}_{t},k,\bs_{t})$ is a Gaussian and benefits from the reparameterization trick using this transformation $\ba^{k-1}_{t} = f_\theta(\epsilon^{k-1}_t;\ba^{k}_{t},k,\bs_{t})$. Thus the gradient of the actor loss can be approximated as
\begin{equation}
\begin{aligned}
    \nabla_\theta \cc{L}(\theta) = &-\nabla_{\ba^0_t}Q_\phi(\bs_t,\ba^0_t)\frac{\partial \ba^0_t}{\partial\theta} \\
    & + \sum_{k=1}^{K} \nabla_{\ba^{k-1}_{t}} \log p_\theta(\ba^{k-1}_{t}|\ba^{k}_{t},k,\bs_{t}) \\
    & \quad \cdot \nabla_\theta f_\theta (\epsilon^k_t;\ba^{k}_t,k,\bs_{t})
\end{aligned}
\label{sac_policy_loss}
\end{equation}
where the term $\frac{\partial \ba^0_t}{\partial\theta}$ is also computed with reparameterization trick as already used in previous methods as direct policy optimization \cite{wang2022diffusion,kang2024efficient}.

\subsection{Hybrid Diffusion Policy}
\label{hydo_update}
To address the challenge of non-prehensile manipulation, we build on the formulation presented in Subsection~\ref{sac-diffusion} by extending it to a hybrid setting, where the policy must handle both discrete and continuous actions. To achieve this, we first augment the HACMan \cite{ZhouJYPH23} objective with entropy regularization terms for both location and motion policies:
\[J_i(\theta) = -Q_\phi(f_i,\ba^m_i) + \alpha \log \pi_{\theta,i}(x_i,\ba^m_i|\bs),\] where $\log \pi_{\theta,i}(x_i, \ba^m_i|\bs)$ includes both location entropy $\log \pi^{\rm{loc}}_i(x_i|\bs)$ and motion parameter's entropy $\log \pi^{\rm{m}}_i(\ba^m_i|\bs)$. As a result, the total objective of the actor is $ J_\theta(\theta) = \sum_{i}^N \ploc (x_i|\bs) J_i(\theta)$.
Similarly, the critic loss 
is defined in Eq.\ref{critic-loss} with an addition of the maximum entropy term to the target 
\begin{align*}
    y_t = r_t + \gamma \mathbb{E}_{x_i \sim \ploc, \ba^m_i \sim \pmot} \big [ &Q_\phi(f_i(\bs_{t+1}, \ba^m_i)) \\
    &- \alpha \log \pi_{\theta,i}(x_i, \ba^m_i|\bs) \big ].
\end{align*}
\begin{algorithm}[htbp]
    \caption{HyDo: Hybrid Diffusion Policy}
    \label{algo:hydo}
    \begin{algorithmic}[1] 
    \State Initialize policy $\pi_\theta$ and critic networks $Q_{\phi_1}$ and 
    $Q_{\phi_2}$
    \State Initialize the target networks: $Q_{\phi'_1}$ and $Q_{\phi'_2}$
    \State Initialize replay buffer: $\cc{D}=\emptyset$
    \While{not converge}
        \State Forward the encoder to compute features $f=f(\bs_t)$
        \State Sample action map $\ba^m_t$ from diffusion policy as in Eq.~\ref{diffusion_policy}, e.g. DDPM or CM sampling. 
        \State Compute Q-value map $Q=Q_\phi(f,\ba^m_t)$
        \State Select contact point $x_i$ using location policy Eq.~\ref{loc_policy}
        \State Select corresponding action's motion parameter $\ba^m_{t,i}$       
        \State Execute action $(x_i, \ba^{m,0}_{t,i})$, observe $r_t$ and $\bs_{t+1}$
        \State Add $\{\bs_t, (x_i, \ba^{m,0}_{t,i}), r_t, \bs_{t+1}\}$ to replay buffer $\cc{D}$
        \State Sample a minibatch $\{\bs,(x_i, \ba^0_{t,i}),r_t,\bs'\}$ from $\cc{D}$
        \State Critic update as with target $y_t$ defined in \ref{hydo_update}.
        \State Actor update with loss $J_\pi(\theta)$ as defined in \ref{hydo_update}
        \State Adjust temperature $\alpha$
        \State Update the target networks like SAC.
    \EndWhile
    \State \textbf{return} final policy $\pi_\theta$.
    \end{algorithmic}
\end{algorithm}

We then introduce Hybrid Diffusion Policy (\acrshort{hydo}), which models the motion parameter policy $\pi_m$ using diffusion models. In particular, diffusion policy $\pi_m$ predicts action map $\ba^m$ as a denoising process
\begin{align*}
    \pi^m(\ba^m|\bs) &= p_\theta(\ba^{m,0:K}|\bs) \\
    &= \mathcal{N}(\bm{a}^{m,K};\bm{0},\bm{I}) \prod_{k=1}^K p_\theta(\ba^{m,k-1}|\ba^{m,k},\bs),
\end{align*}
where we denote $\ba^{m,k}$ is an action map at denoising step $k$. As a result of applying \acrshort{hydo} in \ref{sac-diffusion}, the per-point loss of the actor is rewritten as

\begin{align*}
    J_i(\theta) = &-Q_\phi (f_i, \ba^{m,0}) \nonumber + \alpha_1 \log \ploc_{i}(x_i|\bs) \nonumber \\
    &+ \alpha_2 \sum_{k=0}^K \log p_\theta(\ba^{k-1}_t|\ba^{k}_t,k,\bs),
\end{align*}
for $i=1,\ldots,N$.
Thus, the total objective of the actor is $ J_\pi(\theta) = \sum_{i}^N \ploc (x_i|\bs) J_i(\theta)$, and its gradient is computed using the chain rule through the softmax of $Q$ of the location policy in Eq.~\ref{loc_policy} and the gradient in Eq.~\ref{sac_policy_loss}. Finally, the critic is updated using a standard update in Eq.~\ref{critic-loss} with the following entropy maximization term in the target as
\begin{align}
    y_t = &r_t + \gamma \mathbb{E}\Bigg[ Q_\phi(f_i(\bs_{t+1}, \ba^{m,0}_i)) \nonumber - \alpha \log \ploc_{i}(x_i|\bs_{t+1}) \nonumber \\
    &\quad \nonumber - \alpha \sum_{k=1}^K \log p_\theta(\ba^{k-1}_t|\ba^{k}_t,k,\bs_{t+1}) \Bigg],
\end{align}
where the expectation is taken over both the location and motion policies,  ${x_i \sim \ploc(\cdot|\bs_{t+1}), \ba^{m,0:K}_i \sim \pmot(\cdot|\bs_{t+1})}$.

In addition, the diffusion policy can be replaced by a consistency model $\pi^m(\ba^m|\bs)=\texttt{Consistency\_Model}(\bs;f_\theta)$ without changing the underlying optimization procedure.
We name this variant, Hybrid Diffusion Policy with Consistency Models (\acrshort{hydo} $+$ CM). Both variants are summarized with pseudo-code in Alg.~\ref{algo:hydo}.

\section{Experiments}


In this section, we evaluate \acrshort{hydo} and \acrshort{hydo} $+$ CM, and compare them with the main baseline \acrshort{hacman} and its variants, \acrshort{hacman} $+$ Diff, and \acrshort{hacman} $+$ CM. In addition, we also investigate \acrshort{hydo} (w/o diffusion).
All methods are evaluated on a set of simulated and real-world tasks. Our primary goal is to assess their performance in terms of success rate, behavior diversity, and generalization ability across different task settings. We use the same training settings as HACMan \cite{ZhouJYPH23}. The input is a 4D point cloud obtained by concatenating 3D goal flow vectors with a 1D segmentation mask which indicates if the point belongs to the target object or the background. Segmentation masks are obtained from ground-truth object masks (simulation) or by using background subtraction (real world robot experiments).

\subsection{Experimental Setup}

\begin{figure}[htbp]
    \centering
    \subfloat[Lego]{%
        \includegraphics[width=0.18\columnwidth,height=0.18\columnwidth]{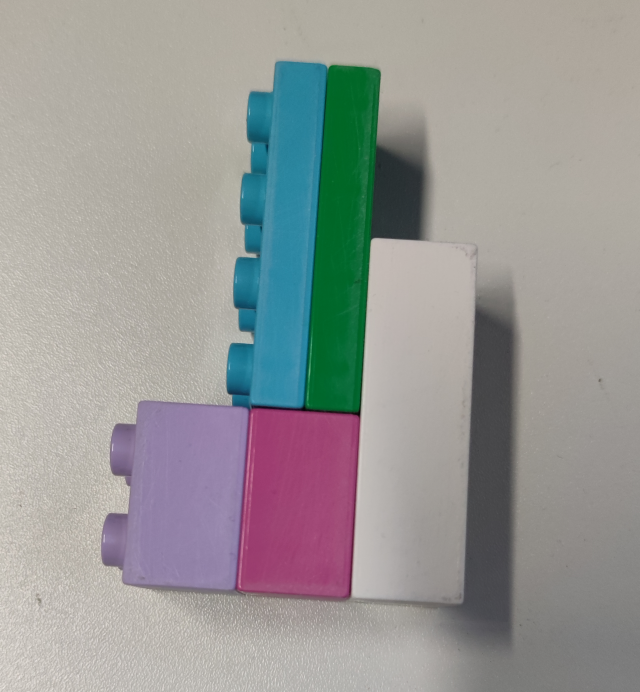}%
        \label{fig:sub1}
    }\hfill
    \subfloat[Lotion]{%
        \includegraphics[width=0.18\columnwidth,height=0.18\columnwidth]{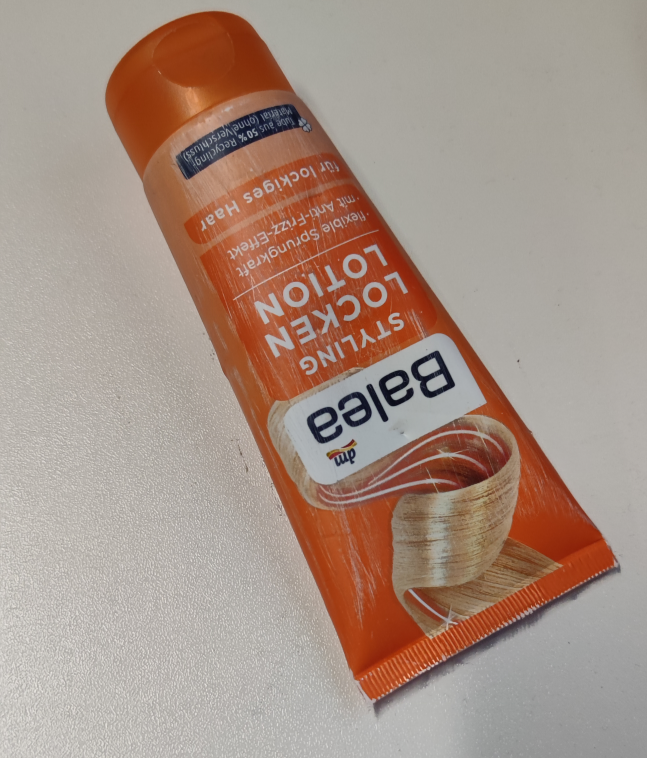}%
        \label{fig:sub2}
    }\hfill
    \subfloat[Milk]{%
        \includegraphics[width=0.18\columnwidth,height=0.18\columnwidth]{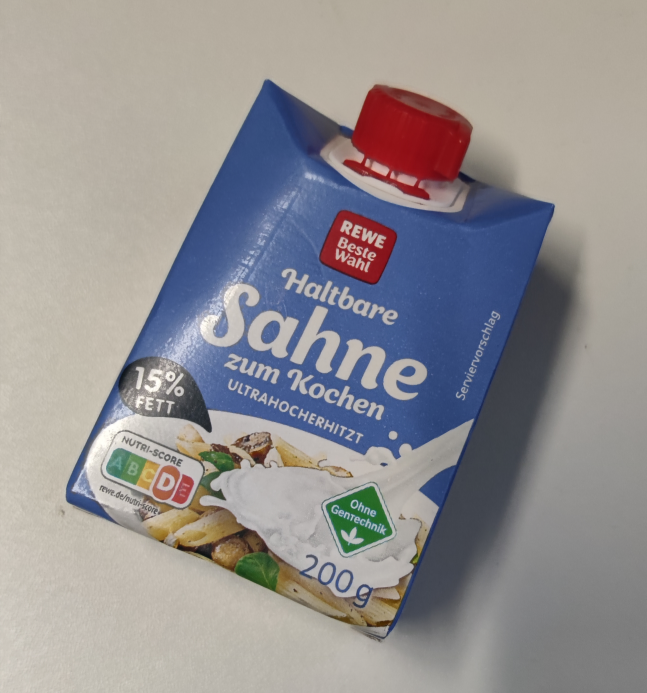}%
        \label{fig:sub3}
    }\hfill
    \subfloat[Soja]{%
        \includegraphics[width=0.18\columnwidth,height=0.18\columnwidth]{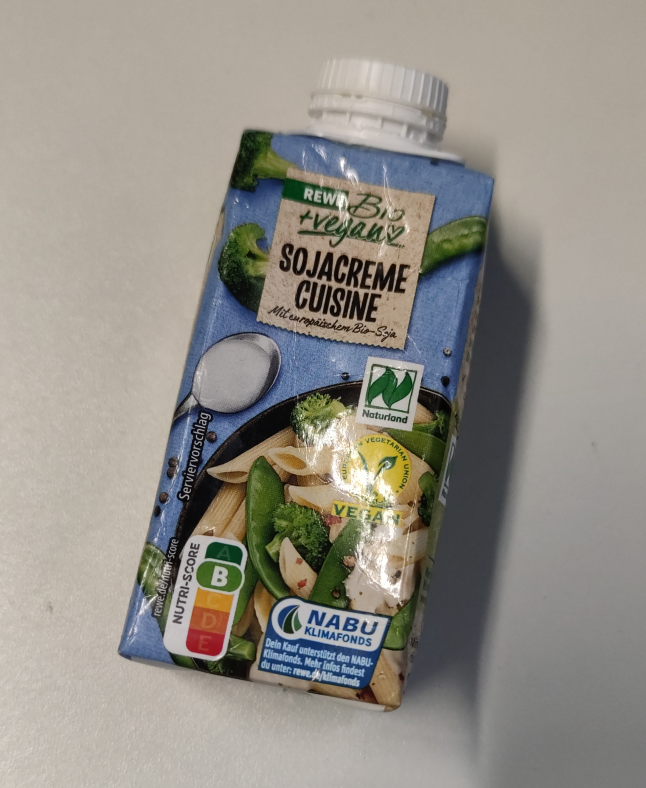}%
        \label{fig:sub4}
    }\hfill
    \subfloat[Cube]{%
        \includegraphics[width=0.18\columnwidth,height=0.18\columnwidth]{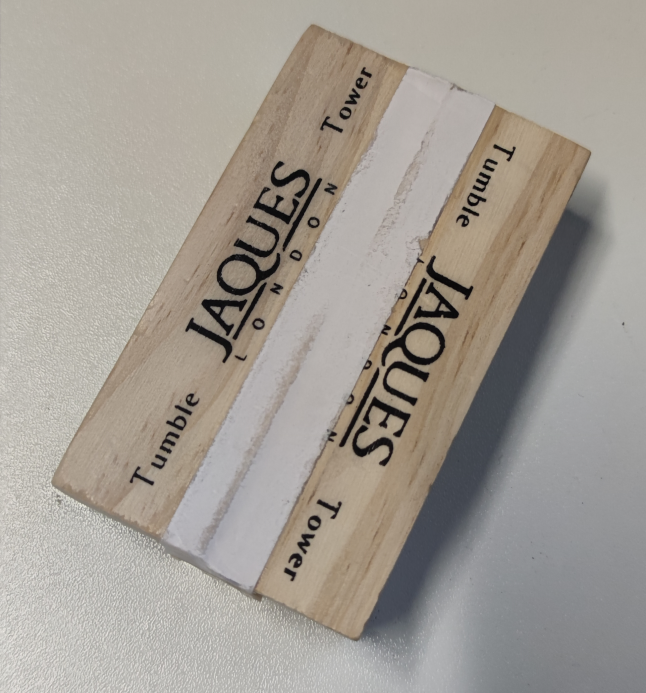}%
        \label{fig:sub5}
    }
    \caption{Set of five objects used for real robot evaluations.}
    \label{fig:eval-objects}
\end{figure}

\begin{figure*}[t]
    \centering
    \begin{minipage}{0.19\textwidth}
            \centering
            \includegraphics[width=\textwidth]{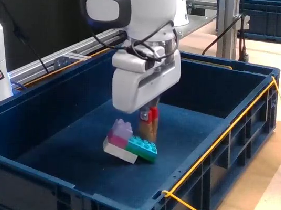}
    \end{minipage}
    \begin{minipage}{0.19\textwidth}
            \centering
            \includegraphics[width=\textwidth]{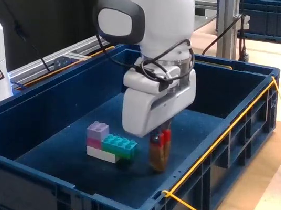}
    \end{minipage}
    \begin{minipage}{0.19\textwidth}
            \centering
            \includegraphics[width=\textwidth]{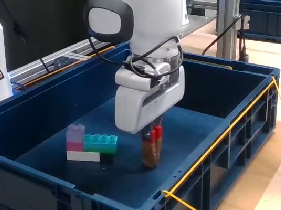}
    \end{minipage}
    \begin{minipage}{0.19\textwidth}
            \centering
            \includegraphics[width=\textwidth]{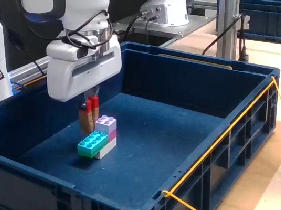}
    \end{minipage}
    \begin{minipage}{0.19\textwidth}
            \centering
            \includegraphics[width=\textwidth]{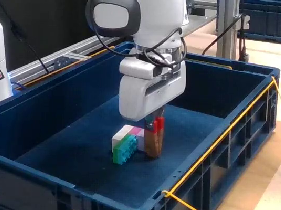}
    \end{minipage}

    \vspace{0.1cm}
    
    \begin{minipage}{0.19\textwidth}
            \centering
            \includegraphics[width=\textwidth]{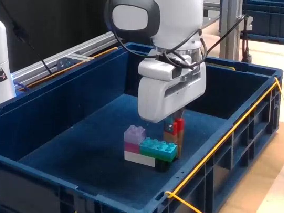}
    \end{minipage}
    \begin{minipage}{0.19\textwidth}
            \centering
            \includegraphics[width=\textwidth]{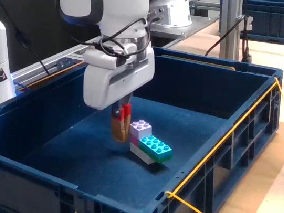}
    \end{minipage}
    \begin{minipage}{0.19\textwidth}
            \centering
            \includegraphics[width=\textwidth]{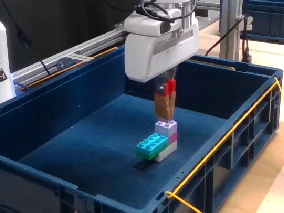}
    \end{minipage}
    \begin{minipage}{0.19\textwidth}
            \centering
            \includegraphics[width=\textwidth]{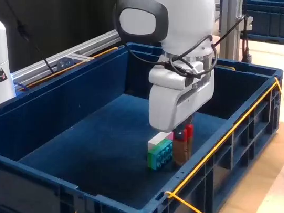}
    \end{minipage}
    \begin{minipage}{0.19\textwidth}
            \centering
            \includegraphics[width=\textwidth]{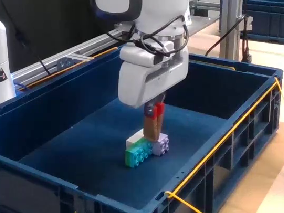}
    \end{minipage}

    \caption{
        A real robot task showcases the multi-modalities of action sequences,  \textbf{(top)} Push$\rightarrow$ Push $\rightarrow$ Push$\rightarrow$ Push$\rightarrow$ Flip; \textbf{(bottom)} Push$\rightarrow$ Push $\rightarrow$ Push$\rightarrow$ Flip$\rightarrow$ Push. In this task, we fixed the goal and initial pose and generated the action sequences with two different random seeds.
    }
    \label{fig:real_trajectory}
\end{figure*}


We validate our method through comparisons and ablation studies performed on the 6D object pose alignment task and translation task introduced in HACMan \cite{ZhouJYPH23}. This task requires diverse non-prehensile manipulations like pushing and flipping to achieve a specified goal pose. {\bf Simulation setting}: The environment is built using Robosuite \cite{zhu2020robosuite} and MuJoCo \cite{todorov2012mujoco} and provides 44 different objects for alignment. The dataset is split into training (32 objects), unseen instances (7 objects), and unseen categories (5 objects). Success is defined by a mean distance of less than 3 cm between the corresponding points of the object and the goal. More details can be found in HACMan \cite{ZhouJYPH23}. {\bf Real robot setting}: The robot system for sim2real evaluations is equipped with a 7DoF Franka Panda arm and three static Realsense cameras. {\bf Evaluations}: We evaluate the algorithms on training objects set. Subsequently, we test these two different configurations: i) Planar goals, where the object starts at a fixed pose and with a randomized planar translation goal pose; and ii) 6D goals, where both the initial and goal poses are stable SE(3) poses that are randomized. Specifically, for both real-world and simulation evaluations, the goals are sampled from SE(3). Initially, object poses are sampled from SE(3) in the air above the center of the bin. The object is dropped into the bin and allowed to settle into a stable position, which is recorded as the goal pose. In simulation, 100 stable poses are collected for each object, and at the start of each episode, a goal is randomly selected from these poses. The location of the selected stable pose is then randomized within the bin.

\subsection{Experimental Results}

\subsubsection{Simulation Results}

Tab.~\ref{tab:simulation_unseen} presents the results of the experiments conducted in simulation for 6D pose tasks on unseen category, unseen instance objects, as well as the objects in the training set. In this evaluation, we report the interquartile mean (IQM) \cite{agarwal2021deep} success rate over 10 seeds and using the best checkpoint for each method.


\begin{table}[ht]
    \centering
    \small
    \caption{Generalization performance on the 6D pose simulation task (using IQM with 95\% confidence intervals).}
    \label{tab:simulation_unseen}
    \resizebox{\columnwidth}{!}{ 
        \begin{tabular}{lccc}
            \toprule
            \textbf{Method} & \textbf{Unseen Category} & \textbf{Unseen Instance} & \textbf{Training Objects}  \\    
            \midrule
            HACMan & $0.760 \pm 0.042$ & $0.818 \pm 0.049$ & $0.769 \pm 0.062$ \\        
            HACMan $+$ Diff & $0.728 \pm 0.041$ & $0.780 \pm 0.028$ & $0.712 \pm 0.047$ \\
            HACMan $+$ CM & $0.671 \pm 0.123$ & $0.703 \pm 0.130$ & $0.632 \pm 0.128$ \\
            \midrule
            HyDo (w/o Diff) & $0.816 \pm 0.026$ & $0.848 \pm 0.032$ & $0.794 \pm 0.028$ \\
            HyDo & $\mathbf{0.843 \pm 0.043}$ & $\mathbf{0.884 \pm 0.046}$ & $\mathbf{0.814 \pm 0.044}$ \\
            HyDo $+$ CM & $0.827 \pm 0.034$ & $0.861 \pm 0.030$ & $0.794 \pm 0.038$ \\
            \bottomrule
        \end{tabular}
    }
\end{table}


Overall, these results indicate a substantial performance improvement for methods which use an entropy term for regularization.
Specifically, HyDo and its variants consistently outperform the HACMan baselines on the \emph{Unseen Category} and \emph{Unseen Instance} object sets. The difference in performance is particularly highlighted in diffusion-based policies ($+$ Diff or $+$ CM), where HyDo and HyDo $+$ CM with the additional entropy regularization terms yields up to a 10\% and 15\% improvement compared to HACMan $+$ Diff and HACMan $+$ CM, respectively. In addition, either with or without the entropy terms, the diffusion policies achieve slightly better performance than consistency models across all evaluation sets.

These findings demonstrate the effectiveness of entropy regularization and underscore its critical role in optimizing complex policies, such as diffusion and consistency models, for improved generalization in unseen scenarios.

\subsubsection{Real Robot Results}

We follow the same setup, e.g., pose randomization, and success criteria as described in HACMan using a set of objects (see Fig.~\ref{fig:eval-objects}) to evaluate the ``All Objects + 6D Goals" on the real-world robot setup, using the trained policies in simulation.



\begin{table}[ht]
    \centering
    \small
    \caption{Success rates for real robot experiments on \\ 
    Planar (left) and 6D (right) goals.}
    \label{tab:real_robot}
    \resizebox{\columnwidth}{!}{ 
        \begin{tabular}{lcccc}
            \toprule
            \textbf{Object} & \textbf{\acrshort{hacman}} & \textbf{\acrshort{hydo} (w/o Diff)} & \textbf{\acrshort{hydo}}  & \textbf{\acrshort{hydo} $+$ CM} \\
            \midrule
            Lego    & 6/10 \quad 6/10 & 8/10 \quad 5/10 &  8/10 \quad 6/10 & 9/10 \quad 7/10 \\
            Lotion  & 6/10 \quad 5/10 & 7/10 \quad 6/10 & 8/10 \quad 7/10 & 7/10 \quad 7/10 \\
            Milk    & 4/10 \quad 6/10 & 8/10 \quad 6/10 & 7/10 \quad 7/10 & 8/10 \quad  7/10 \\
            Soja    & 5/10 \quad  5/10 & 6/10 \quad 4/10 & 6/10  \quad 6/10 & 7/10 \quad 5/10 \\
            Cube    & 5/10 \quad 5/10 & 8/10 \quad 6/10 &  8/10 \quad 5/10 & 9/10 \quad 6/10 \\
            \midrule
            \textbf{Total}   & 26/50 \quad 27/50 & 37/50 \quad 27/50 & 37/50  \quad 31/50 & {\bf 40/50} \quad {\bf 32/50} \\
            \bottomrule
        \end{tabular}
    }
\end{table}

Tab.~\ref{tab:real_robot} reports the evaluation results on both the planar and 6D tasks. We run each method with ten  trials and randomize the initial and target object poses. We keep the initial and target poses similar across evaluations for all methods. The average success rates of the four methods are 53\% for \acrshort{hacman}, 64\% for \acrshort{hydo} (w/o Diffusion), 68\% for \acrshort{hydo}, and 72\% for \acrshort{hydo} $+$ CM. While these results demonstrate that all methods are capable of generalizing to unseen objects in real world scenarios, we do observe a gap between non-diffusion and diffusion policy methods. Compared to an entirely simulated evaluation, this gap is larger if the corresponding method is trained in simulation and applied to real world scenes.
Our hypothesis is that the diverse behaviors of the learned multi-modal diffusion policies allow for better generalization to such new environments. To further assess this diversity, we evaluate multi-modal action sequences under the same environmental conditions, differing only by action sampling seeds.  Fig.~\ref{fig:real_trajectory} shows two distinct action sequences achieving the same goal pose in real robot experiments. This demonstrates how the order of flips and pushes can vary through a diverse execution of actions while it still achieves the same outcome.



\vspace{-0.13cm}
\subsection{Qualitative Policy Diversity Analysis}

\begin{figure}[htbp]
    \vspace{0.1cm}
    \centering
    \begin{minipage}{0.42\columnwidth}
        \includegraphics[width=\textwidth]{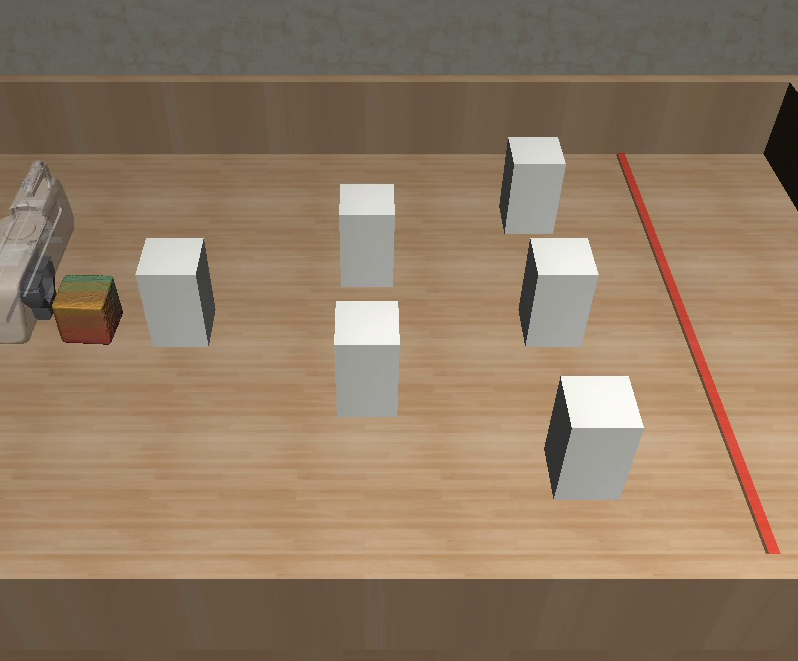}
    \end{minipage}
    \hfill
    \begin{minipage}{0.53\columnwidth}
        \small
        \begin{tabular}{lc}
        \toprule
        \textbf{Method} & \textbf{Entropy} \\
        \midrule
        HACMan & 0.000 \\
        HACMan + Diff & 0.432  \\
        HACMan + CM  & 0.462 \\
        \midrule
        HyDo (w/o Diff) & 0.000 \\
        HyDo     & 0.598  \\
        HyDo + CM     & 0.461 \\
        \bottomrule
        \end{tabular}
    \end{minipage}
    \caption{\rebuttal{\textbf{Left: } A pushing task with $24$ behavioral modes. \textbf{Right: } Behavior entropy of different methodologies.}}
    \label{fig:ent_pushing_task}
\end{figure}

\begin{figure}[htbp]
    \centering
    \begin{minipage}{0.42\columnwidth}
        \includegraphics[width=\textwidth]{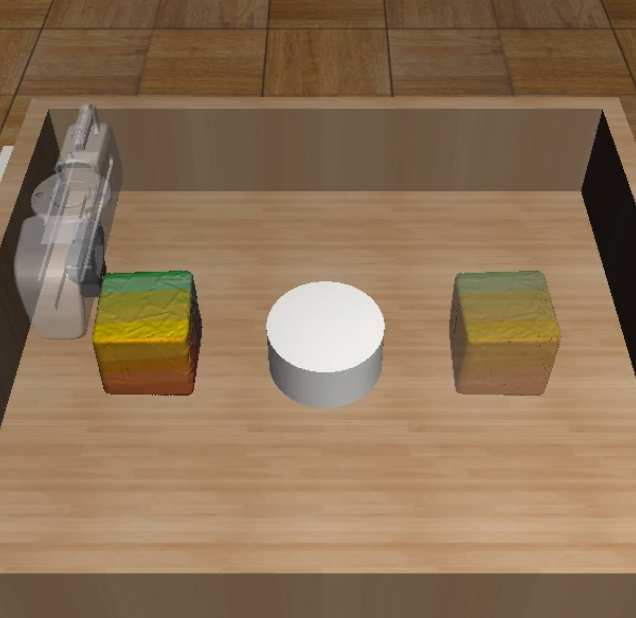}
        \label{diversity_task}
    \end{minipage}
    \hfill
    \begin{minipage}{0.53\columnwidth}
        \includegraphics[width=\textwidth]{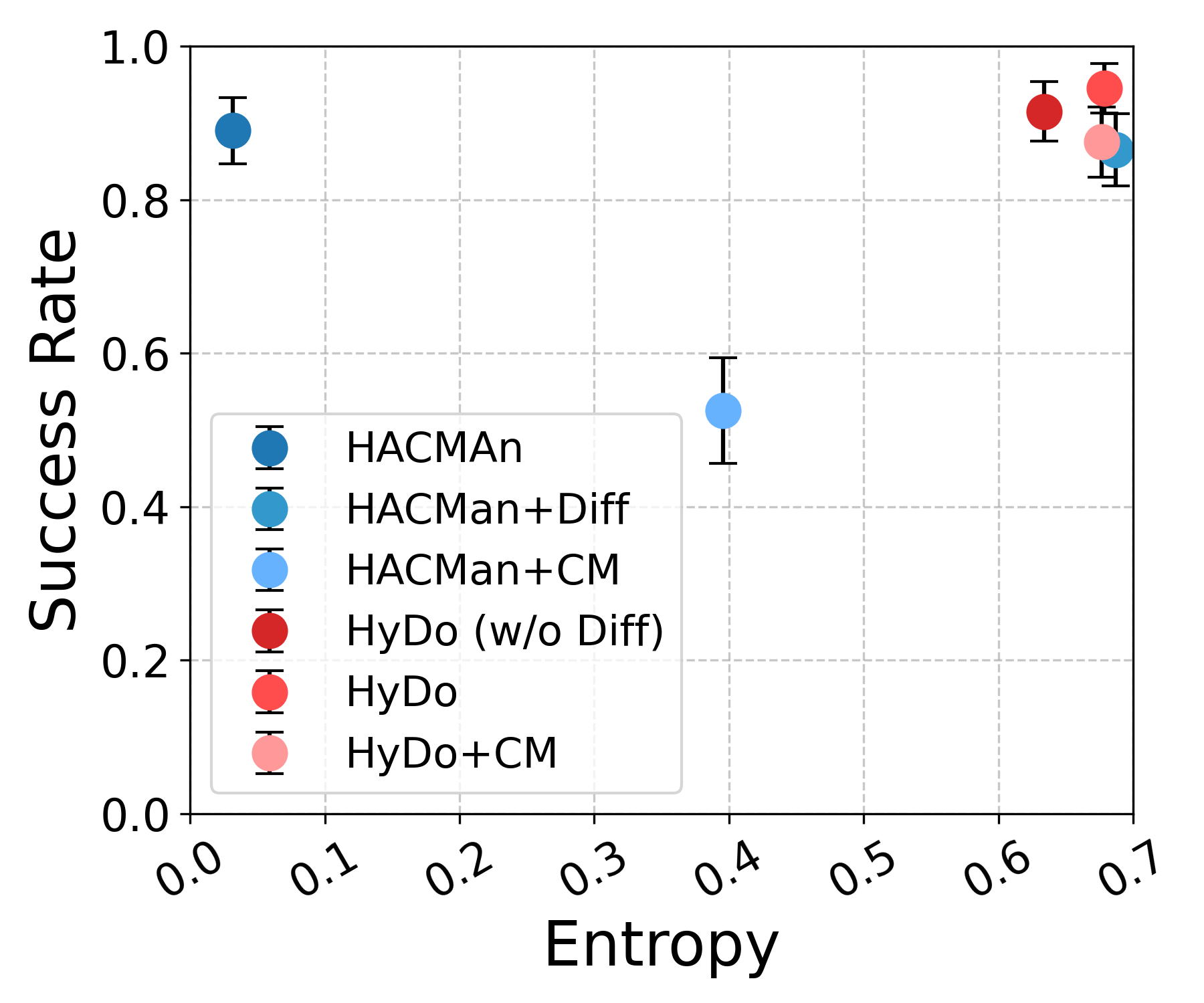}
    \end{minipage}
    \caption{\rebuttal{\textbf{Left: } A push and align task with $2$ behavioral modes. \textbf{Right: } Pareto plot between Entropy and Success Rates.}}
    \label{fig:ent_pushing_task_2}
\end{figure}

Due to the intractability of the diffusion policy's entropy $H(\pi_\theta)$, we follow (Jia et al. 2024) \cite{jia2024towards} to analyze policy diversity via behavior entropy, defined as

\begin{align*}
\cc{H}(\pi(\beta)) = -\sum_{\beta \in \cc{B}} \pi(\beta) \log_{|\cc{B}|} \pi(\beta),
\end{align*}


where $\cc{B}$ is a set of task-specific behavior descriptors. \rebuttal{We first evaluate our method on a task similar to \cite{jia2024towards}, where the robot pushes a cube across a red line (see Fig.~\ref{fig:ent_pushing_task}, left), with $|\cc{B}| = 24$ denoting the number of possible behaviors. We also introduce a new task where the robot must push and align the cube to a target pose by following either the upper or lower trajectory while avoiding white obstacles (see Fig.~\ref{fig:ent_pushing_task_2}, left), resulting in $|\cc{B}| = 2$.}


To enhance diversity during training, we randomly initialize the cube's pose and train all methods until convergence. In the evaluation phase, we fix the initial pose (illustrated in Fig.~\ref{fig:ent_pushing_task} and Fig.~\ref{fig:ent_pushing_task_2}), and perform 500 and 200 trials for pushing and the push-and-align tasks, respectively. In addition, the entropy term for the location actions is disabled to highlight the effect of the entropy on the continuous motion parameters.



\begin{figure}[htbp]
    \centering
    \includegraphics[width=0.5\columnwidth]{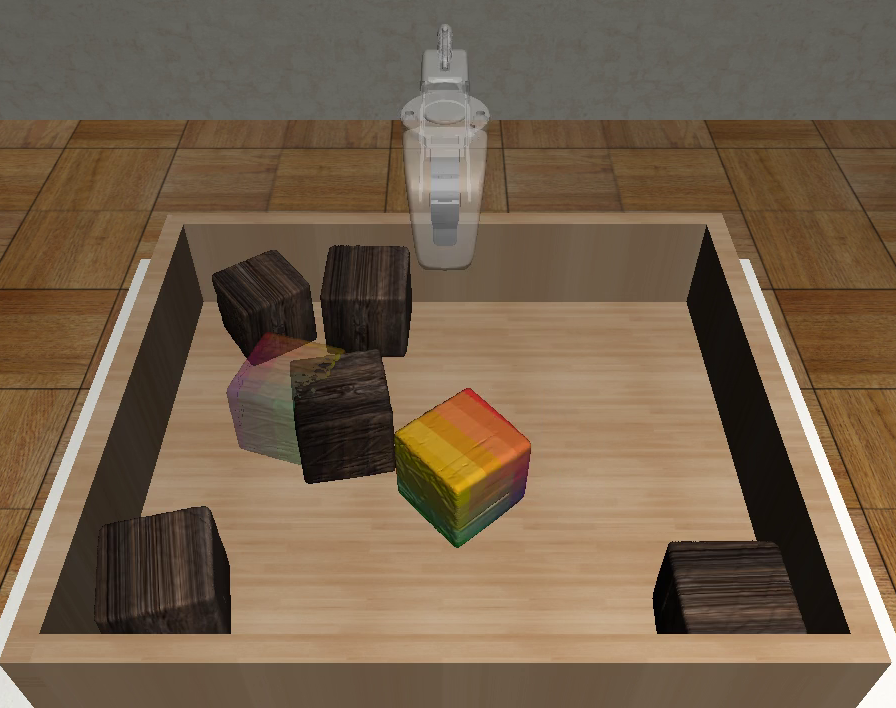}
    \caption{Cluttered environment. The manipulated object is shown in color, the obscured object is the target, and the scene objects are depicted in dark wood.}
    \label{fig:clutter_scene}
\end{figure}

\rebuttal{In the pushing task, HACMan and HyDo (w/o Diff) fail to explore alternative strategies and converge to a single solution (Fig.~\ref{fig:ent_pushing_task}, right), while diffusion-based methods solve the task using diverse trajectories.} For the push-and-align task (Fig.~\ref{fig:ent_pushing_task_2}, right), methods without entropy regularization (e.g., \acrshort{hacman} and \acrshort{hacman} $+$ CM) show reduced behavior diversity (low entropy) and/or lower success rates.

Next, we introduce the cluttered environment featuring varying obstacle objects (represented as a scene point cloud), as shown in Fig.~\ref{fig:clutter_scene}. Table~\ref{tab:clutter_scene} presents the results, indicating that HyDo and HyDo $+$ CM outperform the baselines.

\begin{table}[htbp]
    \centering
    \small
    \caption{Performance under different cluttered scenes.}
    \resizebox{\columnwidth}{!}{ 
    \begin{tabular}{lcccc}
        \toprule
        \textbf{Method} & \multicolumn{3}{c}{\textbf{Numbers of Scene Objects}} \\
        \cmidrule(lr){2-4}
        & \textbf{1} & \textbf{2} & \textbf{5} \\
        \midrule
        HACMan         & $0.527 \pm 0.212$ & $0.495 \pm 0.155$ & $0.421 \pm 0.176$ \\
        HyDo (w/o Diff) & $0.486 \pm 0.121$ & $0.457 \pm 0.113$ & $0.350 \pm 0.110$ \\
        HyDo           & $0.601 \pm 0.100$ & $\mathbf{0.592 \pm 0.091}$ & $\mathbf{0.499 \pm 0.080}$ \\
        HyDo + CM      & $\mathbf{0.615 \pm 0.046}$ & $0.585 \pm 0.059$ & $0.488 \pm 0.042$ \\
        \bottomrule
    \end{tabular}
    }
    \label{tab:clutter_scene}
\end{table}



\subsection{Ablation Study of Consistency and Diffusion Models}
\label{sec:diff_vs_consistency}

To assess the computational speed of consistency and diffusion models with varying denoising steps, we run an experiment measuring inference times for $K \in \{5, 10, 20, 50\}$, using a single-object "Hammer" training setup. As shown in Tab.~\ref{tab:computation_time}, the consistency model \cite{song2023consistency} achieves similar performance to the diffusion model with fewer denoising steps, demonstrating its efficiency advantage. Both models reach a performance plateau around $K=5$, with further increases in $K=50$ leading to a drop in performance.

\begin{table}[h!]
\centering
\small
\caption{Inference Time in milliseconds per sample \\ for each method and diffusion step.}
    \resizebox{\columnwidth}{!}{ 
        \begin{tabular}{lccc}
        \toprule
        \textbf{Method} & $K$ & \textbf{Inference Time (in ms)} & \textbf{Success Rate} \\
        \midrule
        HyDo & 5 & 7.90 $\pm$ 1.06 & 0.684 $\pm$ 0.038\\
        HyDo & 10 & 17.01 $\pm$ 2.66 & 0.646 $\pm$ 0.059 \\
        HyDo & 20 & 29.27 $\pm$ 3.67 & 0.677 $\pm$ 0.045 \\
        HyDo & 50 & 74.23 $\pm$ 10.25 & 0.644 $\pm$ 0.025 \\
        \midrule
        HyDo + CM & 5 & 7.51 $\pm$ 1.14 & 0.787 $\pm$ 0.077\\
        HyDo + CM & 10 & 14.89 $\pm$ 1.77 & 0.731 $\pm$ 0.007 \\
        HyDo + CM & 20 & 26.72 $\pm$ 3.65 & 0.713 $\pm$ 0.058 \\
        HyDo + CM & 50 & 67.70 $\pm$ 9.93 & 0.575 $\pm$ 0.009 \\
        \bottomrule
        \label{tab:computation_time} 
        \end{tabular}
    }
    \vspace{-0.45cm}
\end{table}

\section{CONCLUSIONS}

We presented Hybrid Diffusion Policy (\acrshort{hydo}), an online diffusion-based off-policy maximum entropy RL algorithm for 6D non-prehensile manipulation. We derived a principled objective, i.e. the maximum entropy regularization, that considers diffusion policies as a class of stochastic policies. We showed that treating the stochastic diffusion policy with a principled objective significantly improves its performance in RL applications. Our qualitative results indicated that online RL is hard for learning multi-modal policy distributions with diffusion models, as it can make diffusion policies converge to uni-modal quickly. Therefore, stochastic diffusion-based and entropy maximizing RL algorithms can be a promising combination for improved exploration strategies and learning more diversity behaviors. For future work, we envision extending our approach to more complex, dynamic environments, such as closed-loop settings and tasks requiring continuous adaptation \cite{cho2024corn}.


%

\appendices





\ifCLASSOPTIONcaptionsoff
  \newpage
\fi



\bibliography{root}{}
\bibliographystyle{IEEEtran}

\end{document}